\documentclass{article}
\usepackage[T1]{fontenc}
\usepackage[utf8]{inputenc}
\usepackage{array}
\usepackage{float}
\usepackage{multirow}
\usepackage{amsmath}
\usepackage{amssymb}
\usepackage{stmaryrd}
\usepackage{graphicx}
\usepackage[authoryear]{natbib}
\usepackage{microtype}
\usepackage[unicode=true,
 bookmarks=false,
 breaklinks=false,pdfborder={0 0 1},backref=section,colorlinks=false]
 {hyperref}
\hypersetup{pdftitle={A template for the arxiv style},
 pdfauthor={David S.~Hippocampus, Elias D.~Striatum},
 pdfsubject={q-bio.NC, q-bio.QM},
 pdfkeywords={First keyword, Second keyword, More}}

\makeatletter

\providecommand{\tabularnewline}{\\}

\usepackage{arxiv}

\usepackage{url}
\usepackage{amsfonts}
\usepackage{nicefrac}
\usepackage{lipsum}
\usepackage{graphicx}
\usepackage{doi}

\usepackage{float} 
\usepackage{graphicx}
\usepackage{microtype}	
\usepackage{tabularx}

\usepackage{tikz}
\usepackage{comment}
\usepackage{color}

\usepackage{amssymb}
\usepackage{array}
\usepackage{tabularx}

\usepackage{amsmath}
\usepackage{amsthm}
\usepackage{booktabs}

\usepackage{algorithm}
\usepackage{algorithmic}

\title{A template for the \emph{arxiv} style}

\author{%
	Thao Minh Le\textsuperscript{1}, 
	Vuong Le\textsuperscript{2}, 
	Kien Do\textsuperscript{1}, 
	Sunil Gupta\textsuperscript{1}, 
	Svetha Venkatesh\textsuperscript{1}, 
	Truyen Tran\textsuperscript{1} \\[2ex]
	\textsuperscript{1}Applied Artificial Intelligence Institute, Deakin University, Australia\\[1ex]
	\textsuperscript{2}Amazon, Melbourne, Australia\\[2ex]
	\texttt{\{thao.le,k.do,sunil.gupta,svetha.venkatesh,truyen.tran\}@deakin.edu.au}\\[1ex]
	\texttt{levuong@amazon.com}
}

\date{}

\makeatother

\begin{document}
	\title{Finding the Trigger: Causal Abductive Reasoning on Video Events}
	
	\maketitle
	
	
	\global\long\def\TaskName{\text{CARVE}}%
	\global\long\def\Dataset{\text{CARVE}}%
	\global\long\def\ModelName{\text{CERN}}%
	\global\long\def\eg{e.g.}%
	\global\long\def\DatasetReal{\text{EpicKitchen-AR}}%
	
	\begin{abstract}
		\inputencoding{latin9}This paper introduces a new problem, Causal
Abductive Reasoning on Video Events (CARVE), which involves identifying
causal relationships between events in a video and generating hypotheses
about causal chains that account for the occurrence of a target event.
To facilitate research in this direction, we create two new benchmark
datasets with both synthetic and realistic videos, accompanied by
trigger-target labels generated through a novel counterfactual synthesis
approach. To explore the challenge of solving CARVE, we present a
Causal Event Relation Network (CERN) that examines the relationships
between video events in temporal and semantic spaces to efficiently
determine the root-cause trigger events. Through extensive experiments,
we demonstrate the critical roles of event relational representation
learning and interaction modeling in solving video causal reasoning
challenges. The introduction of the CARVE task, along with the accompanying
datasets and the CERN framework, will advance future research on video
causal reasoning and significantly facilitate various applications,
including video surveillance, root-cause analysis and movie content
management.
 
	\end{abstract}
	
	\section{Introduction}
	
	\inputencoding{latin9}Modern AI research has achieved unprecedented
progress in exploring statistical patterns from massive amounts of
data to infer knowledge about the world. This is often referred to
as inductive reasoning. For many inductive reasoning tasks, AI systems
have already outperformed humans, such as image recognition \cite{he2015delving}
and voice recognition \cite{assael2016lipnet}. However, while humans
excel at generating hypotheses and intuitions about the \emph{causes
of an observation,} this capability remains a blind spot for AI systems.
For example, current medical AI systems are trailing behind human
doctor in determining the probable cause of a disease diagnosis from
patient's medical history, lifestyle and their recent travel history.
This type of inference is called \emph{Abductive reasoning} and was
first introduced by Charles Sanders Peirce \cite{peirce1974collected}.
In the field of video analysis, this fundamental problem translates
into important and impactful requirement of finding the cause-effect
pairs of events within the video.

\begin{figure*}
\begin{centering}
\includegraphics[width=0.95\textwidth]{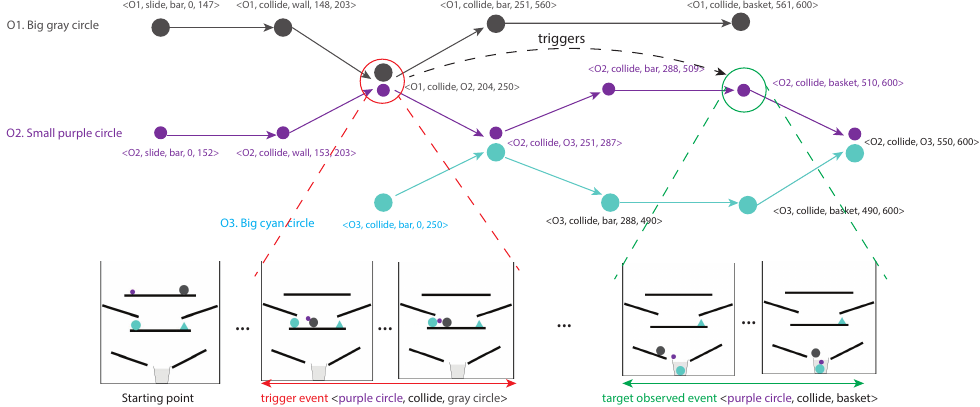} 
\par\end{centering}
\vspace{-2mm}
 \caption{Illustration of event chains and trigger-target event pairs in $\protect\Dataset$
task and example in the accompanying dataset. Videos are generated
by a 2D physics simulator using predefined visual scenes. Video events,
\emph{\textless object\_id\_1, object\_id\_2, interaction\_type,
start\_time, end\_time\textgreater}, are defined as interactions
between a dynamic object and an object partner within a time interval.
Each dynamic object creates a chain of events (3 in this example,
marked with corresponding colors). These chains are merged by mutual
events, resulting in a graph of events. Explanation and target event
pairs are identified by comparing the original video and counterfactual
videos. Best viewed in color.\label{fig:Examples-of-abductive-pairs}}
\end{figure*}

Aiming at this problem, this paper studies a new task called \emph{Causal
Abductive Reasoning on Video Events }($\TaskName$) requiring AI systems
to comprehend the causal relations embedded in the videos at the \emph{event
level} and generate hypotheses about the \emph{trigger event -} a
prior event in the video that most probably caused the occurrence
of\emph{ }a\emph{ }query\emph{ target event}. $\TaskName$ promises
significant breakthrough in numerous applications. For instance, in
video surveillance, $\TaskName$ can help trace back the causal chain
of a suspicious behavior to see why it was anomalous; in sports analysis,
the task supports analyzing an unexpected move of a player on the
field and suggest the likely reason such as a change in strategy or
an injury; in movie production, the storyline of long videos can be
broken down to identify illogical or redundant parts and provide suggestions
for edition.

Along with proposing the $\TaskName$ task, we introduce two benchmark
datasets built with counterfactual synthesis through interventional
schemes. The first one is the namesake $\Dataset$ dataset featuring
noise-free videos of dynamic objects that are created through a physics
simulator. The trigger-target labels are generated by counterfactual
video generation. The second dataset \emph{EpicKitchen-AR} leverages
the realistic videos from action forecasting task in the EpicKitchen
dataset and augment them with abductive events labels, also through
counterfactual synthesis. Figure \ref{fig:Examples-of-abductive-pairs}
illustrates the $\TaskName$ task and an example of the datasets.

Attempting at solving the new challenge, we propose a novel \emph{Causal
Event Relation Networks ($\ModelName$)}. The method focuses on modeling
both temporal dependencies and semantic dependencies between video
events toward finding their trigger-target relationships. $\ModelName$
has this done by building a temporal-constrained directed event graph
network built on input event features. The framework is generic to
a variety of event feature representations demonstrated with its performance
on both synthetic $\Dataset$ and the lifelike EpicKitchen-AR data. 

Extensive experiments on the two datasets indicate that the task introduces
a new distinctive challenge to video understanding field, demonstrated
on the fact that fine-tuned large-scale video recognition models,
such as MViTv2 \cite{li2022mvitv2}, and powerful large video-language
models such as Video-LLaVA \cite{lin2023video} all fall short in this
new task, due to their inability to disentangle the event causal relations
with other types. The experiments also show the advantages of the
new proposed method $\ModelName$ against various baseline methods
and suggest promising directions for the task.

Overall, this work aims at initiating a research topic of causal abductive
reasoning for video events and makes three contributions: (1) Proposing
a new task $\TaskName$ and building two rich and diverse datasets
using counterfactual synthesis (2) A novel neural framework for event-graph
modeling toward effectively solving $\TaskName$ task and (3) Extensive
experiments that reveal the challenges of the task and indicate directions
for development in this new front.

	\section{Related Work \label{sec:Related-Work}}
	
	\inputencoding{latin9}\textbf{Visual Abductive (VA) reasoning}: This
task is traditionally formulated as question-answering \cite{cherian20222,dang2021hierarchical,yi2020clevrer,gao2023mist}where
the cause of an event is selected from multiple choices annotated
by human. Another line of works approach abductive reasoning in the
form of linguistic explanations \cite{liang2022visual,zhao2022videoabc,hessel2022abduction,suchan2018visual,li2020causal}
with strong reliance on textual captioning. The concrete trigger-target
pair settings make $\TaskName$ distinctive from these work that aims
at generating a generic hypothetical explanation that only might be
associated with an actually happened event. Furthermore, this new
task requires the focus on spatio-temporal visual properties and avoid
linguistic bias and ambiguity, therefore making it closer to the targeted
applications. Our work also transcend from visual reasoning about
object's relations and physics \cite{ding2021dynamic,ates2020craft,locatello2020object,girdhar2019cater,yi2020clevrer}
by extending to life-like video events. 

\textbf{Visual Abductive datasets}: Currently available VA datasets,
including VAR \cite{liang2022visual}, CLEVRER family \cite{yi2020clevrer,mao2022clevrer}
are all based on artificial explanation generated by either human
annotations which are subjective to human annotators or preset written
scripts which are biased to linguistic commonsense. To aim at more
genuine visual casual data, we break this tradition with a new counterfactual
synthesis procedure to reconstruct the objective cause-effect facts. 

\textbf{Event-relation modeling:} Previous works focus on event localization
\cite{liu2021multi,dai2017temporal,xia2022learning} and video future
events predictions \cite{lei2020more,park2020visualcomet}. Structured
representations of events are popular for these problem with action
scene graphs \cite{ji2020action} and action role labels \cite{sadhu2021visual}.
Event-level graph in our work are built on top of the recent progress
in video spatio-temporal graph models such as unified video-text event
graph \cite{ayyubi2022multimodal} and pseudo-3D scene graph \cite{cherian20222}.
The event-related applications recently saw further progress with
vision transformer-based methods \cite{arnab2021vivit,liu2022video,li2022mvitv2},
however, video causal reasoning remains a challenge, especially at
event level. 

\textbf{Generic abductive reasoning}: Outside of computer vision,
abductive reasoning has been explored using event calculus and temporal
action logic \cite{denecker1992temporal}. In linguistics, abduction
for text-based events in commonsense narrative contexts is collected
and labeled by human labors \cite{bhagavatula2019abductive}. It is
also explored in medical diagnosis \cite{elsenbroich2006case} and
root-cause-analysis \cite{schoenfisch2018root}. These works give
insight and ideas to the new abductive reasoning task using causal
relation of video events introduced in this paper.

	\section{Causal Abductive Reasoning on Video Events\label{sec:Task-Def}}

\subsection{Task Definition}

\inputencoding{latin9}The \textbf{C}ausal \textbf{A}bductive \textbf{R}easoning
on \textbf{V}ideo \textbf{E}vents \textbf{($\TaskName$) }involves
analyzing a sequence of chronological events to identify preceding
events that contributed causally to the occurrence of a target event.
Here, we consider $\TaskName$ in the video understanding context.
Specifically, within a sequence of events found in a video, we consider
a \textbf{\emph{target}}\textbf{ event} at position $N$: $e^{\text{\text{target}}}=e_{N}$.
The $\TaskName$ asks AI systems to find all the grouth-truth trigger
events $E^{trigger}$ within the set of premise events $E^{\text{premise}}=\left\{ e_{1},...,e_{N-1}\right\} $
that are the causal reasons of the target. This setting excludes the
case where $E^{trigger}=\varnothing$, meaning the target is an exogenous
event triggered by factors outside of the video. We treat our $\TaskName$
as a binary classification problem where $y_{i}=1$ indicates that
a preceding event $e_{i}$ is a cause of $e^{\text{target}}$ and
$y_{i}=0$ otherwise. Formally, our task is to maximize the probability
that an event $e_{i}\in E^{\text{trigger}}$ is correctly identified
as a trigger event for the target event:\vspace{-2mm}

\begin{equation}
\theta^{*}\hspace{-0.15em}=\hspace{-0.15em}\text{argmax}_{\theta}\hspace{-0.15em}\sum_{e_{i}\in E^{\text{premise}}}\hspace{-0.15em}P_{\theta}\hspace{-0.15em}\left(y=1\hspace{-0.15em}\mid\hspace{-0.15em}e_{i},e^{\text{target}},E^{\text{premise}}\right),\label{eq:task_definition}
\end{equation}
where $\theta$ indicates model parameters. 

This task is different from the previous visual abductive reasoning
tasks by raising the requirement that the \emph{causing factor must
be grounded into a subset of premise events} instead of as generic
textual description of the cause either in free form \cite{liang2022visual}
or QA \cite{yi2020clevrer,mao2022clevrer} formats. This requirement
bring the task closer to a large set of applications where the root
cause needs to be traced back to previous video events.

\subsection{Counterfactual Causal Event Datasets}

The top challenge for observational models for causal discovery tasks
like CARVE is to separate causal relations with associative biases.
For example, ``eating ice cream'' is usually associated with ``sun
burnt'' while actually caused by ``hot weather''. This separation
theoretically requires training and evaluation data \emph{with ground
truth cause-effect pairs discovered with interventional experiments}
\cite{pearl2000models}. Traditionally, these interventions are estimated
by randomized controlled trials but they are universally costly and
mostly infeasible for life-like video events. This challenge in building
causal data encouraged a popular alternative direction where the event
causes are specified as \emph{linguistic explanations} learned from
causal facts extracted from literature corpus \cite{liang2022visual}.
Another work around is through human annotation, either in explanation
form or in question-answering format \cite{yi2020clevrer,mao2022clevrer}.
While avoiding the roadblock, these datasets are prone to diverging
from the true causal relation between events, often leaning on linguistic
causal commonsense provided by explanatory generative models - which
are overly smooth or human annotator who can be objective and inconsistent.
For example, while ``eating ice cream'' is commonly explained by
``hot weather'', in a particular video, it can be actually caused
by the real event of ``attending a birthday party''.

To break through these limitations, we propose to directly address
interventional data generation by using counterfactual synthesis where
the candidate trigger events are one-by-one omitted from the video
to observe its consequential role to the target event. This approach
helps establish two video event datasets as benchmark for the $\TaskName$
task: The $\mathbf{\TaskName}$ dataset produced using a fully controlled
2D physics simulator, and the realistic \emph{EpicKitchen-AR} leverages
the real-world scenarios using life-like videos extracted from the
popular EpicKitchen-100 dataset. Meanwhile, the \emph{EpicKitchen-AR}
set ensures the practicality in dealing with lifelike videos.

\subsubsection{CARVE dataset}

We aim at building a large scale dataset with clean videos and clear
causal relationships that can be used to validate the hypotheses and
verify core operation of methods. To generate the videos, we utilize
the 2D physics simulator provided by \cite{ates2020craft} to generate
complex dynamic physical interactions of 2D objects. We generate 10K
videos with over 250K abductive event pairs of target and trigger
events. The videos and their abductive event pairs are divided into
train, validation and test split with a ratio of 60:20:20.

\textbf{Video generation: }The videos are 10 seconds long of $256\times256$
pixel resolutions. They are created from 20 pre-defined scenes with
physical interactions of \emph{dynamic objects }and \emph{static scene
elements. }There are 48 variants of dynamic objects derived by selecting
among 2 sizes (small, large), 3 shapes (cube, triangle, circle) and
8 colors. We design 7 static scene elements\emph{ }(ramp, platform,
button, basket, left wall, right wall and ground). Object and element
positions are randomly initiated.

\textbf{Dynamic object events:} In this dataset, we specify video
events based on the interactions of objects, named \emph{dynamic-object
events}. We first track the movement of each object and detect its
\emph{state changes} caused by \emph{interactions} with other objects
by looking for sudden changes in velocity or direction of its movement.
We then match the pairs of state changes of two objects happening
within the proximal time and space to locate the interactions. The
\emph{interaction} is classified into two types of \emph{collision
}and \emph{slide}. For interactions involving more than two objects,
we break them into different combinations of two-object interactions. 

The event is specified by its start and end times indicating \emph{the}
\emph{time interval between two consecutive collisions that result
in a change in the main object's movement direction}. The event feature
vector is then form as: \emph{\textless object\_id\_1, object\_id\_2,
interaction\_type, start\_time, end\_time}\textgreater . 

To form the relational structure of the events, we build a directed
graph $\mathcal{G}=(\mathcal{V},\mathcal{E})$ of dynamic-object events,
where the nodes $\mathcal{V}$ represents events and the edges $\mathcal{\mathcal{E}}$
represent the temporal order between them. Within the graph $\mathcal{G}$,
for each object, we obtain a set of \emph{chain of events} through
that object's lifeline throughout the course of a video. These event
chains intersect at the events that involve two dynamic objects. See
Figure~\ref{fig:Examples-of-abductive-pairs} for an illustration
.

\begin{algorithm}[t] 
\small 
\caption{Abductive Event Pairs Extraction} 	
	\hspace*{\algorithmicindent} \textbf{Input}:   
	$\mathcal{G}$: Event graph of the input video \\ \hspace*{\algorithmicindent} 
	\hspace*{\algorithmicindent} \hspace*{\algorithmicindent} $e^{\text{target}}_t$: target event of main object $t$ \\ 
	\hspace*{\algorithmicindent} \hspace*{\algorithmicindent} \hspace*{\algorithmicindent} $C$: counterfactual dynamic objects of $e^{\text{target}}_t$ \newline         
	\hspace*{\algorithmicindent} \textbf{Output}: $H$: List of trigger events for $e^{\text{target}}_t$

	\begin{algorithmic}[1] 	
		\STATE Initialize $H \gets \{\}$ 	
		\FOR {\textbf{each} $c \in C$} 		
			\STATE $e^{\textrm{first}}_{c} \leftarrow$ {first event in the event sequence in $\mathcal{G}$ associated with $c$} 	
			\STATE $paths \leftarrow$ depth\_first\_search\_all\_paths($\mathcal{G}, e^{\textrm{first}}_c \rightarrow e^{\textrm{target}}_t$)
			\IF {$paths \equiv \emptyset$}                 
				\STATE continue             
			\ENDIF 	
			
			\FOR {\textbf{each} $path \in paths$}	 			
				\STATE $h \leftarrow$ {first event of interactions of $c$ and a dynamic object partner} 		
				\IF {$h \equiv \emptyset$}                      
					\STATE $h \leftarrow$ first in the event chain associated with $c$                 
				\ENDIF 			
				\IF {$h \not\in H$}
                     \STATE add $h$ to $H$
                \ENDIF
			\ENDFOR
   	\ENDFOR   
	\end{algorithmic} 
\label{algo:abductive_pairs_extraction} 
\end{algorithm} 

\textbf{Generating Trigger-target event pairs: }

For each original video, we generate a \emph{counterfactual variant}
by removing a dynamic object at the beginning of the original video
and re-running simulation. We then pick one event from the event graph
of the original video as a target event; and then try to search for
the existence of the target event in a counterfactual video graph
by similar object attributes and time-space location. If the target
event is not found in the counterfactual video, the removed dynamic
object is marked as \emph{affecting} the event. In contrast, if no
match is found, we say the removed object to be \emph{non-affectting}
of the target event. 

For each affecting object of the target event, we identify the first
event involving a dynamic object partner in its event chain as a possible
trigger event. This means\emph{ a single target event may have multiple
trigger events, depending on the number of affecting objects involved}.
If no event involving a dynamic object partner exists, the first event
in the chain is chosen (See Algorithm~\ref{algo:abductive_pairs_extraction}).
This procedure is repeated for all possible target events of the original
video graph. We note that target and trigger events do not necessarily
share a common dynamic object. Instead, the target event can be an
end result of a series of collisions via multiple intermediate objects,
which is initiated by the trigger event. 

\begin{figure}
\begin{centering}
\includegraphics[width=0.6\columnwidth]{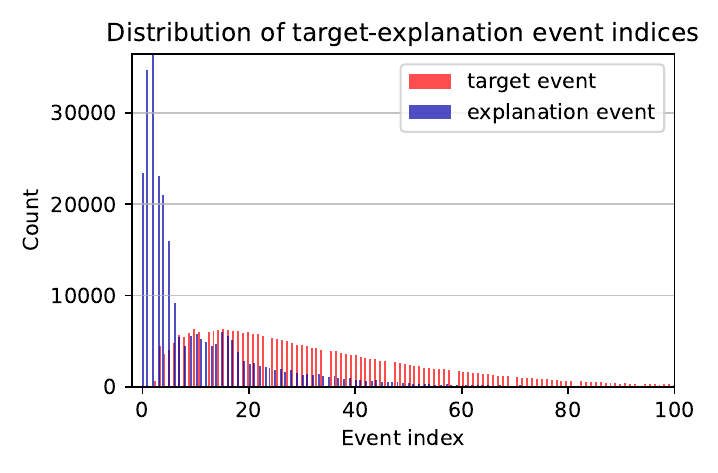} 
\par\end{centering}
\vspace{-4mm}
\caption{Distribution of target and trigger event locations in time in $\protect\Dataset$.
Only 100 first events are visible.\label{fig:Distribution-of-event-ids}}
\end{figure}

\textbf{Data splits and statistics:} After discarding 9 videos where
no abductive event pairs, the remaining videos are divided into 3
splits:\emph{ train}, \emph{validation} and \emph{test}, with a ratio
\emph{60:20:20}. This results in a large-scale dataset of a total
254,278 pairs where the train, validation and test split take 152,916
pairs, 50,822 pairs and 50,540 pairs, respectively.

Figure~\ref{fig:Distribution-of-event-ids} shows the distributions
of target and trigger event locations in our $\Dataset$ dataset.
This graph shows that the trigger events are neither simply the first
event in a chain nor the event right before the target event, instead
they can be any preoccuring events. This quality of data prevent ML
models from tactically remembering heuristics and requires them to
properly discover the event causal relations. 

\subsubsection{The EpicKitchen-AR Dataset}

While $\Dataset$ is clean with strong trigger-target relations, we
would like to extend the benchmark to \emph{real-world videos} where
these relations are naturally entangled with other types of relations
and environmental factors. To this end, we employ the Epic-Kitchen-100
\cite{damen2022rescaling}, a major dataset originally designed for
action forecasting task. This task involves predicting an action starting
at a particular time based on a sequence of past event observations. 

To upgrade Epic-Kitchen to an Abductive reasoning dataset, we need
to generate a set of trigger-target event pairs for the video. This
is a big challenge because unlike $\Dataset$, we cannot create counterfactual
videos to determine the trigger event for the target. To overcome
this challenge, we leverage the current state-of-the-art action anticipation
method, AFFT \cite{zhong2023anticipative}, as an oracle model to
generate pseudo-counterfactual trigger-target pairs.  

The process starts by running the action prediction oracle on all
videos and selecting ones where the oracle model produces correct
top-5 action predictions. Next, we join adjacent actions with the
same label into events. We then consider the target event to be the
event that contains the predicted action. The pseudo-counterfactual
process is done by masking out each past event in the observation
sequence one-by-one to observe changes in the oracle prediction of
the target event. Now with these hypothetical event sequences, similarly
to $\Dataset$ dataset, we define a trigger event to be the first
event whose ablation causes a false prediction of the target event.
These trigger-target event pairs constitute our EpicKitchen-AR dataset
consisting of 1,259 abductive event pairs, each found in one distinct
video clip. We then use 1,000 videos for training and the remaining
259 videos for validation.

	\section{Causal Event Relation Networks \label{sec:Method}}
	
\begin{figure}
	\begin{centering}
		\includegraphics[width=0.75\columnwidth]{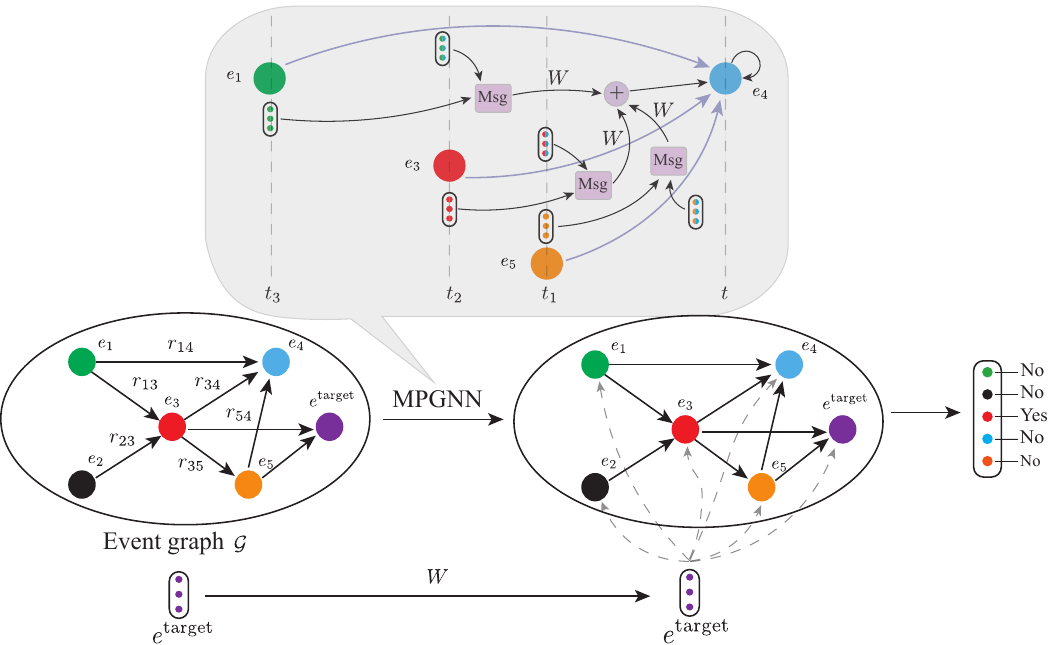} 
		\par\end{centering}
	\vspace{-2mm}
	\caption{Overview of $\protect\ModelName$. Given a target event $e^{\text{target}}$
		and its premise events, we build a directed graph of events $\mathcal{G}$
		based on their temporal distance. Edge feature vectors $r_{ij}$ (arrows)
		represent multi-aspect relations between events. We use a novel message
		passing scheme (Msg) to refine events in consideration of their surrounding
		events. Gray box illustrates how event $e_{4}$ receives information
		from preceding events $e_{1},e_{3},e_{5}$ to refine its representation.
		Refined event features are then bound with the target event and eventually
		mapped into scores for label prediction.\label{fig:Overview-method}}
\end{figure}

	\inputencoding{latin9}In this section, we attempt to solve the $\TaskName$
by devising \emph{Causal event relation network} ($\ModelName$),
a neural framework that analyzes event temporal-semantic graph to
quantify the trigger-target relationships. The overall architecture
is demonstrated in Figure~\ref{fig:Overview-method}.

\subsection{Event Representation\label{subsec:Event-Representation}}

We aim at designing a generic model that supports a wide variety of
event representation vectors $e_{i}$. The particular process to generate
$e_{i}$ is up to the characteristics of the dataset. For the $\TaskName$
dataset, we define an event as the interaction of two objects within
a time interval between two collisions. For an event $e_{i}$, we
denote $x_{m}$ and $x_{p}$ to be the aggregated dynamic attributes
(color, shape, position etc.) of the main and partner objects during
the course of the event; and $ts$ and $te$ be the start and end
times. They stack up to form the event representation vector $e_{i}$
: 

\begin{equation}
e_{i}=<[x_{m},x_{p}],\left[ts,te\right]>.\label{eq:event_representation}
\end{equation}
Differently, for EpicKitchen-AR, event representation $e_{i}$ is
a composition vector including spatio-temporal visual features of
the segments extracted using the TSN networks \cite{zhong2023anticipative},
embeddings of the action label using DistillBERT \cite{sanh2019distilbert},
and event timestamps.

\subsection{Temporal-Semantic Event Graphs\label{subsec:Object-centric-Event-Graph}}

Given the event sets $\mathcal{V}$, our objective is to model their
causal relationship through building a graph-based representation
multi-relational graph $\mathcal{G}=\left(\mathcal{V},\mathcal{E}=\left\{ \mathcal{E}^{\text{temp}},\mathcal{E}^{\text{sem}}\right\} \right)$.
Here the edges $\mathcal{E}$ capture not only their temporal relations
$\mathcal{E}^{\text{temp}}$ but also their semantic relations $\mathcal{E}^{\text{sem}}$.
These two types of edges are formed in the process described in this
section.

\textbf{Event temporal relations} $\mathcal{E}^{\text{temp}}$: Considering
two events $e_{i}$ and $e_{j}$ as defined by Eq.~(\ref{eq:event_representation})
from the event set $\mathcal{V}$, their temporal relationship $r_{ij}^{\text{temp}}$
is determined based on their order in time. This can be quantified
using their temporal distance $d\left(e_{i},e_{j}\right)$ \cite{zhang2013modeling}
which is calculated solely based on their start and end times. Assuming
that the event $e_{i}$ occurs during time interval $\left[ts_{i},te_{i}\right]$
and event $e_{p}$ during $\left[ts_{j},te_{j}\right]$, their temporal
distance is defined by: \vspace{-2mm}

\begin{equation}
\small d\left(e_{i},e_{j}\right)=\left\{ ts_{j}-ts_{i},te_{j}-te_{i},ts_{j}-te_{i},te_{j}-ts_{i}\right\} .\label{eq:temporal_distance}
\end{equation}

We use Allen's interval algebra \cite{allen1983maintaining} to determine
the temporal relations between $e_{i}$ and $e_{p}$ and identify
their order in time. Fundamentally, if $ts_{i}>ts_{j}$ then $e_{i}$
is considered happening before $e_{j}$, denoted as $\left(e_{i},e_{j}\right)_{i<j}$
. If $ts_{i}=ts_{j}$ then end times will be considered. Details on
this order rules are provided in the Supplementary. This temporal
order forms the direction of edges in graph $\mathcal{G}$. We also
use the temporal distance Eq.~(\ref{eq:temporal_distance}) to assign
edge weights. We now have a weighted directed graph $\mathcal{G}(\mathcal{V},\mathcal{E^{\text{temp}}})$
where information flow obeys the time order of video data in which
only prior events can trigger the occurrence of later events but not
in the reversed direction.

\textbf{Event semantic relations }$\mathcal{E}^{\text{sem}}$: In
addition to being temporally related, events are also connected by
their semantic associations. These relations are formed by the input
event representation of dimension \emph{d }that is created accordingly
in each dataset (such as Eq.~(\ref{eq:event_representation}) for
For $\Dataset$ dataset). As event relations are naturally multi-dimensional
event relations (for example ``force'' and ``direction'' in object
interaction), we use a $k$-dimensional vector $r_{e_{i}e_{p}}^{\text{sem}}\in\mathbb{R}^{k}$
to represent the semantic edge features. This is an improvement from
the common practice of studies using scalar pairwise structural relationships
\cite{arnab2021unified,mao2018hierarchical,wang2019zero}. The feature
is formed by a bilinear operator: \vspace{-2mm}

\begin{align}
r_{e_{i}e_{j}}^{\text{sem}} & =\phi\left(e_{i},e_{j}\right)\label{eq:semantic_relation_weights}\\
 & =\gamma_{ij}*\text{tanh}\left(e_{i}^{\top}W^{[1:k]}e_{j}+b\right),
\end{align}
where $\gamma_{ij}\in\left[0,1\right]$ is a distance-based penalty
factor of temporal distance with detailed formulation is provided
in Supp. $W^{[1:k]}\in\mathbb{R}^{d\times d\times k}$ and $b\in\mathbb{R}^{k}$
are learnable parameters. The bilinear operator is used on the basis
of each slice of the tensor $W^{[1:k]}$ ($k\equiv d$ in our implementation).

Finally, we combine the two attributes of event relations to output
a directed graph structure $\mathcal{G}(\mathcal{V},\mathcal{E})$
representation of V nodes of an input video: 

\begin{equation}
r_{ij}=r_{ij}^{\text{temp}}*r_{ij}^{\text{sem}};\mathcal{E}=\left\{ r_{ij}\right\} _{i=1,j=1}^{V}.
\end{equation}

\subsection{Abductive Reasoning on Event graphs}

This section describes the learning-to-reason process on the event
graphs toward finding trigger events for the targets to solve the
task $\TaskName$ in Eq.~(\ref{eq:task_definition}). Given the trigger
event groundtruth $e_{i}$ of a target event $e_{t}$ within its premise
events $E\left(e_{t}\right)=\mathcal{G}\left(\mathcal{V}=\left\{ v_{i}\right\} _{i<t},\mathcal{E}=\left\{ r_{ij}\right\} _{i,j<t}\right)$,
solving Eq.~(\ref{eq:task_definition}) becomes maximizing 
\begin{equation}
P\left(y=1\mid e_{t},e_{i},E\left(e_{t}\right);\theta\right).
\end{equation}
In our framework, we estimate this probability using an inference
procedure using a classifier $f_{\theta}\left(\cdot\right)$ on top
of the embeddings of target event $z_{t}=q_{\theta}\left(e_{t}\right)$
and trigger event candidates $h_{i}=p_{\theta}\left(e_{i},E\left(e_{t}\right)\right)$:

\begin{equation}
P\left(y=1\mid e_{t},e_{i},E\left(e_{t}\right)\right)=f_{\theta}\left(h_{i},z_{t}\right).\label{eq:logistic_classifier}
\end{equation}

In our implementation, classifier $f_{\theta}\left(\cdot\right)$
is a logistic classifier using concatenated $[h_{i},z_{t}]$ as input.
The target embedding function $q_{\theta}\left(e_{t}\right)$ is a
linear transformation. Especially, the candidate representation learning
\textbf{$p_{\theta}\left(e_{i},E\left(e_{t}\right)\right)$} is implemented
as a novel message passing scheme with L layers composed of two phases:

\begin{figure}
\begin{centering}
\includegraphics[width=0.6\columnwidth]{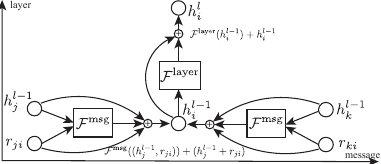} 
\par\end{centering}
 \caption{Message passing with skip connections along \textquotedblleft message\textquotedblright{}
and \textquotedblleft layer\textquotedblright{} axis. $\mathcal{F}^{\text{msg}}$
and $\mathcal{F}^{\text{layer}}$ are non-linear functions. Illustrating
with two neighbors $j$ and $k$ of node $i$.\label{fig:Message-passing-paradigm}}
\end{figure}

\textit{Message aggregation with skip connections:} 
\begin{alignat}{1}
\mathcal{F}_{i}^{\text{layer}}\hspace{-0.15em}\left(\hspace{-0.15em}h_{i}^{l-1}\hspace{-0.15em}\right)\hspace{-0.15em} & =\frac{W_{1}^{l-1}}{\mid N\left(i\right)\mid}\left(\sum_{j\in N\left(i\right)}\text{\ensuremath{\mathcal{F}}}_{j}^{\text{msg}}\left(\left(h_{j}^{l-1},r_{ji}\right)\right)\right)+\nonumber \\
 & \frac{W_{2}^{l-1}}{\mid N\left(i\right)\mid}\hspace{-0.15em}\sum_{j\in N\left(i\right)}\left(\hspace{-0.15em}h_{j}^{l-1}+r_{ji}\hspace{-0.15em}\right),\label{eq:message_skip_connections}
\end{alignat}

\begin{align}
\text{\ensuremath{\mathcal{F}}}_{j}^{\text{msg}}\hspace{-0.15em}\left(\hspace{-0.15em}\left(\hspace{-0.15em}h_{j}^{l-1},r_{ji}\hspace{-0.15em}\right)\hspace{-0.15em}\right) & =\text{ReLU}\hspace{-0.15em}\left(\hspace{-0.15em}W_{3}^{l-1}\hspace{-0.15em}\left[\hspace{-0.15em}h_{j}^{l-1}\hspace{-0.15em},r_{ji}\hspace{-0.15em}\right]\hspace{-0.15em}\right)\hspace{-0.15em};h_{j}^{0}\hspace{-0.15em}=\hspace{-0.15em}e_{i},\label{eq:message_skip_connections_2}
\end{align}

\textit{Layer transformation with skip connections:} 
\begin{align}
h_{i}^{l} & =\text{ReLU}\hspace{-0.15em}\left(\hspace{-0.15em}h_{i}^{l-1}\hspace{-0.15em}+\hspace{-0.15em}W_{4}^{l-1}\mathcal{F}_{i}^{\text{layer}}\left(\hspace{-0.15em}h_{i}^{l-1}\hspace{-0.15em}\right)\hspace{-0.15em}\right),\label{eq:layer_skip_connections}
\end{align}
where $N\left(i\right)$ indicates neighborhood of size $\left|N\left(i\right)\right|$
of node $i$. $W_{1},W_{2},W_{3},W_{4}$ are network parameters.

Unlike the standard message passing, ours uses skip connections during
both message aggregation and layer transformation (Figure~\ref{fig:Message-passing-paradigm}).
These skip connections are of crucial roles: Those along the ``message''
axis offer an alternative path for information to directly flow from
neighbors to node $i$, therefore facilitating information aggregation
within a neighborhood. Meanwhile, the skip connections at the layer
transformation phase create a self-loop information to update a node's
representation by itself at the current layer. As a result, it enables
information from distant events to better reach later events through
$L$ layers using only single-hop neighborhood instead of having to
do costly multi-hop alternatives like in \cite{feng2022powerful}.
Besides increasing the expressive power of the network, these skip
connections also reduce gradient vanishing and facilitate training. 

The network is trained end-to-end using the binary cross-entropy loss
$\mathcal{L}_{\text{BCE}}=-\frac{1}{D}\sum_{i=1}^{D}y_{i}\text{log}\tilde{y}_{i}+\left(1-y_{i}\right)\text{log}\left(1-\tilde{y}_{i}\right)$
where $D$ is the size of the training data and $\tilde{y}_{i}=f\left(h_{i},z_{t}\right)$
indicates model predictions.

\subsection{Modeling Scope and Limitation}

The scope of the $\TaskName$ task and $\ModelName$ model in this
paper limit at endogenous causal relations, meaning that we only consider
causal pairs of events that exist within a single video. This scope
is relevant to vast video analysis applications such as surveillance,
sport video and movie analysis. Extensions can be further made to
the datasets and method to consider the out-of-video exogenous causes
as hypothetical event graph nodes. Another limitation of the EpicKitchen-AR
dataset is that we use an oracle event prediction model to create
counterfactual causal pair. This creates a relative upper bound accuracy
of the labels depending on this oracle, in contrast to the $\Dataset$
dataset where the labels are absolute through actual simulations.

	\section{Experiments\label{sec:Experiments}}

\subsection{Data preparation}

\textbf{$\Dataset$ dataset:} To measure efficiency together with
accuracy, along with the full $\Dataset$ dataset of 10K videos and
250K+ abductive event pairs, we create two smaller subsets with the
same diversity measures. Their sizes are 5K videos/80K pairs and 1K
videos/18K respectively. 

Toward realistic event representation, we only use object features
that can be extracted from visual observation. Specifically, we use
a combined feature of \emph{2D positions}, \emph{velocity} and \emph{one-hot
embeddings} of static object attributes. This results in a feature
vector of 71 dimensions per object, per timestep.

\textbf{$\DatasetReal$ dataset: }As mentioned in Section~\ref{subsec:Event-Representation},
we use spatio-temporal features of video segments as event representations
which are the extracted TSN features following \cite{zhong2023anticipative}.
To enrich the event representations, we also incorporate label embeddings
derived from action label annotations using DistillBERT \cite{sanh2019distilbert}. 

\inputencoding{latin9}\begin{table}
\begin{centering}
\begin{tabular}{>{\centering}p{1.8cm}>{\centering}p{0.6cm}>{\centering}p{0.6cm}>{\centering}p{0.6cm}|>{\centering}p{0.6cm}>{\centering}p{0.6cm}>{\centering}p{0.6cm}}
\hline 
\multirow{2}{1.8cm}{{\footnotesize{}Model}} & \multicolumn{3}{c|}{{\footnotesize{}Val. Accuracy (\%)}} & \multicolumn{3}{c}{{\footnotesize{}Test Accuracy (\%)}}\tabularnewline
\cline{2-7} \cline{3-7} \cline{4-7} \cline{5-7} \cline{6-7} \cline{7-7} 
 & {\footnotesize{}1K}  & {\footnotesize{}5K}  & {\footnotesize{}10K}  & {\footnotesize{}1K}  & {\footnotesize{}5K}  & {\footnotesize{}10K}\tabularnewline
\hline 
\hline 
{\footnotesize{}Rand. guess}  & {\footnotesize{}0.60}  & {\footnotesize{}0.60}  & {\footnotesize{}0.60}  & {\footnotesize{}0.60}  & {\footnotesize{}0.60}  & {\footnotesize{}0.60}\tabularnewline
\hline 
{\footnotesize{}First collision}  & {\footnotesize{}8.72}  & {\footnotesize{}8.72}  & {\footnotesize{}8.72}  & {\footnotesize{}9.30}  & {\footnotesize{}9.30}  & {\footnotesize{}9.30}\tabularnewline
\hline 
{\footnotesize{}Node embs}  & {\footnotesize{}18.92}  & {\footnotesize{}25.66}  & {\footnotesize{}32.36}  & {\footnotesize{}18.08}  & {\footnotesize{}27.65}  & {\footnotesize{}32.62}\tabularnewline
\hline 
{\footnotesize{}LSTM}  & {\footnotesize{}29.29}  & {\footnotesize{}31.22}  & {\footnotesize{}38.03}  & {\footnotesize{}27.17}  & {\footnotesize{}33.62}  & {\footnotesize{}38.42}\tabularnewline
\hline 
{\footnotesize{}BiLSTM}  & {\footnotesize{}28.21}  & {\footnotesize{}32.06}  & {\footnotesize{}38.02}  & {\footnotesize{}27.42}  & {\footnotesize{}34.15}  & {\footnotesize{}37.90}\tabularnewline
\hline 
{\footnotesize{}Transformer}  & {\footnotesize{}24.36}  & {\footnotesize{}31.73}  & {\footnotesize{}41.28}  & {\footnotesize{}23.29}  & {\footnotesize{}34.24}  & {\footnotesize{}41.75}\tabularnewline
\hline 
{\footnotesize{}$\ModelName$}  & \textbf{\footnotesize{}32.05}{\footnotesize{} } & \textbf{\footnotesize{}37.9}{\footnotesize{} } & \textbf{\footnotesize{}44.38}{\footnotesize{} } & \textbf{\footnotesize{}30.90}{\footnotesize{} } & \textbf{\footnotesize{}40.50}{\footnotesize{} } & \textbf{\footnotesize{}43.86}\tabularnewline
\hline 
\end{tabular}
\par\end{centering}
\vspace{2mm}
\caption{Experiments using object-centric event features on the $\protect\Dataset$
dataset and its subsets. \label{tab:Experimental-object-centric-ev-feat}}
\end{table}

\subsection{Baselines and settings}

To set the high bar for the challenge, we implement a wide set of
baselines ranging from classic methods to popular modern architectures.
They include \emph{first collision}, \emph{direct node embeddings},
\emph{LSTM} \cite{hochreiter1997long}, \emph{BiLSTM} \cite{graves2005framewise},
\emph{Transformer} \cite{vaswani2017attention}, and \emph{Video-LLaVA}
\cite{lin2023video}. The \emph{first collision} baseline always predicts
the first collision in video input as the trigger event. The \emph{node
embeddings} baseline makes prediction of about trigger events solely
based on their features without considering their relations. For sequential
models such as LSTM and BiLSTM, we first sort all the events localized
from an input video by their start times then treat them as input
sequences. To assure fairness, we conduct experiments with the popular
large video-language model Video-LLaVA only on $\DatasetReal$ because
they were trained with only real-world videos and language descriptions,
which are not available in the visual-only $\Dataset$ dataset. See
Supp. for more details on these settings. 

Compared to these baselines is our \emph{Causal event relation network}
$(\ModelName)$ using a $4$-layer message passing framework for experiments
with the $\Dataset$ dataset and 2-layer for the $\DatasetReal$ by
default. These relatively deep graph structure allow the model to
reach information from events living further in the past. Thy are
possible thanks to the skip connections that avoid gradient vanishing
issues.

\subsection{Results on CARVE dataset}

We compare the effectiveness of $\ModelName$ against the baselines
by feeding them the same object-centric event features in Eq. (\ref{eq:event_representation}).
The results (given in Table ~\ref{tab:Experimental-object-centric-ev-feat})
confirm that the task cannot be solved by either making random predictions
or selecting the first collision in video input as the explanation
event. We also observe that BiLSTM does not offer any benefits compared
to LSTM but rather degrades the performance with the backward information
in some cases (37.90\% vs. 38.42\% on test split). This suggests that
for this task, information strictly flows from past events to later
events, verifying our inductive bias used in designing $\ModelName$. 

Among the baselines, Transformer is the closest to our event graph
network since both methods model the pairwise relationships between
events. However, Transformer is less sample efficient hence, struggling
to generalize with data scarcity. Our $\ModelName$ model further
benefits from the explicit use of the temporal orders of events and
their distance in time, clearly outperforming Transformer in these
experiments. This advantage is even more significant in experiments
with limited training data (1K and 5K).

\subsubsection{Ablation studies and analysis: }

We justify the roles of $\ModelName$'s components by a series of
ablation studies (Table~\ref{tab:Ablation-studies}):

\textbf{Replacing event representations:}

\emph{Without object-centric event representations: }When replacing
object-centric features with generic video features using a large-scale
video representation MViTv2 \cite{li2022mvitv2} pretrained on Kinectics-400
\cite{kay2017kinetics}. This is the current SoTA model across different
image and video understanding tasks. These features are also combined
with event timestamps to make sure it is a fair comparison. Table~\ref{tab:Ablation-studies}
shows that using objectless features significantly degrades $\ModelName$'s
performance on $\Dataset$ by over 47.0\%.

\emph{Fine-tuning large video recognition models}: To further verify
the importance of object-centric representation and temporal information
of events in learning, we fine-tune the MViTv2 models (S and B) to
learn the mapping from video features to event features without providing
event timestamps. Table~\ref{tab:Ablation-studies} indicates that
increasing both model size and training data do not enhance the model's
ability to localize events in time. This demonstrates their limitations
in detecting subtle object movements and distinguishing object-centric
motions essential for event representations.

\textbf{Ablating model's components:}

We ablate each relation from the full $\ModelName$ and observe the
impacts on the performance. In details:

\textit{\emph{T}}\emph{emporal relations} $\mathcal{E}^{\text{temp}}$
shows its critical roles when its ablation creates a significant drop
of 38.1\%/36.5\% on val/test split\textit{. }Similarly, only \emph{semantic
relations} $\mathcal{E}^{\text{sem}}$ does not work well alone either,
with a drop of around 9.0\%.

With message passing of 2 layers instead of 4 as default, the performance
dropped by 3.0\%. More importantly, the \textit{Message skip connections}
shows their roles as ablating them has a bigger negative impact on
the performance of around 5.0\%.\textit{ Similarly, }when we remove
the direct connections $h_{i}^{l-1}$, the performance also drops
over 5.0\%. Together, \textit{Ablating all skip connections} (Eqs.~(\ref{eq:message_skip_connections}-\ref{eq:layer_skip_connections}))
results in a combined degradation by around 7.0\%.

\textit{\emph{Finally, when we remove the}}\textit{ distance penalty
in event relations} ($\gamma_{ip}$ factor in Eq.~$(\ref{eq:semantic_relation_weights})$),
we see a major effect of performance drop of 8.9\%/8.6\% on val/test
sets. 

\begin{table}
\begin{centering}
\begin{tabular}{>{\centering}p{1.5cm}>{\centering}p{3.5cm}>{\centering}p{2cm}}
\hline 
\multirow{1}{1.5cm}{\centering{}Experiment} & Component & \multicolumn{1}{c}{{\footnotesize{}Val. Acc (\%)}}\tabularnewline
\hline 
\hline 
\multirow{3}{1.5cm}{{\small{}Replacing event features}} & {\small{}MViTv2-S feats with event timestamps} & {\small{}23.26 ($\downarrow$47.5\%)}\tabularnewline
\cline{2-3} \cline{3-3} 
 & {\small{}MViTv2-S feats w/o event timestamps (finetuned)} & {\small{}5.89 ($\downarrow$86.7\%)}\tabularnewline
\cline{2-3} \cline{3-3} 
 & {\small{}MViTv2-B feats w/o event timestamps (finetuned)} & {\small{}5.65 ($\downarrow$87.2\%)}\tabularnewline
\hline 
\multirow{7}{1.5cm}{{\small{}Ablating model's components}} & {\small{}Only temp. relations } & {\small{}27.46 ($\downarrow$38.1\%) }\tabularnewline
\cline{2-3} \cline{3-3} 
 & {\small{}Only sem. relations } & {\small{}40.27 ($\downarrow$9.3\%) }\tabularnewline
\cline{2-3} \cline{3-3} 
 & {\small{}2-layer msg passing } & {\small{}42.90 ($\downarrow$3.3\%) }\tabularnewline
\cline{2-3} \cline{3-3} 
 & {\small{}W/o msg skip connections } & {\small{}42.05 ($\downarrow$5.2\%) }\tabularnewline
\cline{2-3} \cline{3-3} 
 & {\small{}W/o layer skip connections } & {\small{}41.93 ($\downarrow$5.5\%) }\tabularnewline
\cline{2-3} \cline{3-3} 
 & {\small{}W/o skip connections } & {\small{}41.20 ($\downarrow$7.2\%) }\tabularnewline
\cline{2-3} \cline{3-3} 
 & {\small{}W/o distance penalty } & {\small{}40.41 ($\downarrow$8.9\%) }\tabularnewline
\hline 
{\small{}Full model} &  & \textbf{\small{}44.38}{\small{} }\tabularnewline
\hline 
\end{tabular}
\par\end{centering}
\vspace{2mm}
\caption{Ablation studies and analysis on $\protect\Dataset$ dataset. $\downarrow$
indicates performance drop from the full model.\label{tab:Ablation-studies}}
\end{table}

\begin{table}
	\begin{centering}
		\begin{tabular}{cccc}
			\toprule 
			\multirow{2}{*}{{\footnotesize{}Model}} & \multicolumn{3}{c}{{\footnotesize{}Val. Accuracy (\%) }}\tabularnewline
			\cmidrule{2-4} \cmidrule{3-4} \cmidrule{4-4} 
			& {\footnotesize{}Visual} & {\footnotesize{}Label} & {\footnotesize{}Combined}\tabularnewline
			\midrule
			\midrule 
			{\footnotesize{}{}Rand. guess} & {\footnotesize{}5.56} & {\footnotesize{}5.56} & {\footnotesize{}5.56}\tabularnewline
			\midrule 
			{\footnotesize{}{}Node embs} & {\footnotesize{}6.59} & {\footnotesize{}39.77} & {\footnotesize{}41.47}\tabularnewline
			\midrule 
			{\footnotesize{}{}LSTM} & {\footnotesize{}33.33} & {\footnotesize{}45.17} & {\footnotesize{}45.74}\tabularnewline
			\midrule 
			{\footnotesize{}{}Transformer} & {\footnotesize{}32.56} & {\footnotesize{}45.17} & {\footnotesize{}42.25}\tabularnewline
			\midrule 
			{\footnotesize{}Video-LLaVA (zero-shot)} & {\footnotesize{}NA} & {\footnotesize{}NA} & {\footnotesize{}4.26}\tabularnewline
			\midrule 
			{\footnotesize{}Video-LLaVA (in-context)} & {\footnotesize{}NA} & {\footnotesize{}NA} & {\footnotesize{}34.50}\tabularnewline
			\midrule 
			{\footnotesize{}{}$\ModelName$} & \textbf{\footnotesize{}37.2} & \textbf{\footnotesize{}46.90} & \textbf{\footnotesize{}47.29}\tabularnewline
			\bottomrule
		\end{tabular}
		\par\end{centering}
	\vspace{2mm}
	\caption{Experiments on the $\protect\DatasetReal$ dataset. \label{tab:Exps-abductive-ek100}}
\end{table}

\subsection{Results on EpicKitchen-AR:}

We compare the performance of $\ModelName$ against the baselines
on $\DatasetReal$, examining different types of event representations,
including visual embeddings, label embeddings and combined embeddings
of the features (Table~\ref{tab:Exps-abductive-ek100}). We particularly
compare to the popular multi-purpose large video-language model Video-LLaVA
\cite{lin2023video} where it has access to language descriptions
(event labels), to maximize its capabilities. We tried both zero-shot
and in-context learning settings. For zero-shot, we instruct Video-LLaVA
to select the trigger amongthe set of premise events for a target
event and match its generated responses to ground-truth events. In
the in-context setting, each test sample is accompanied by two examples
with the target-trigger annotations from the train set as in-context
samples. Details of the prompts used are provided in the Supplementary. 

Table \ref{tab:Exps-abductive-ek100} shows that\emph{ $\ModelName$
clearly outperforms the baselines for all of the cases} with visual-only,
label-only and combined visual-label event representations. Interestingly,
the direct node embeddings baseline completely failed to find trigger
events in visual-only case, suggesting event visual representations
are highly complex and requires stronger modeling capability. Most
notably, Video-LLaVA struggled to understand the causal relationships
between events, highlighting the lack of reasoning ability despite
its capacity to generate plausible text descriptions about video contents.
Our proposed method $\ModelName$ also \emph{consistently outperforms
LSTM and Transformer}. It is to note that Transformer gets 3.0\% performance
drop in the combined case compared to label-only. This is explainable
as Transformer is known to be data hungry and could not generalize
with the available data to make visual feature useful.

While setting the state of the art, the performances of $\ModelName$
and other baselines are still far from perfect at the $\TaskName$
task on both datasets, suggesting \textit{an open playground for future
works} in this new task.

	\section{Discussion \label{sec:Discussion}}
	
	\inputencoding{latin9}This paper promotes the task of Causal Abductive
Reasoning on Video Events ($\TaskName$), an important but under-explored
AI capability. The task is supported by two new benchmark datasets
with clean and realistic videos created using our counterfactual synthesis
procedures. We also introduced the Causal Event Relation Networks
($\ModelName)$, a new neural framework operating on event graph can
learn to effectively find the trigger events in videos. While CERN
outperforms many baselines, including finetuned large-scale generic
models, the new task is proved to be highly challenging and calling
for further advances toward causal abductive reasoning for visual
data.

{\small}{\bibliographystyle{plain}
\bibliography{main}

\begin{thebibliography}{48}
\providecommand{\natexlab}[1]{#1}
\providecommand{\url}[1]{\texttt{#1}}
\expandafter\ifx\csname urlstyle\endcsname\relax
  \providecommand{\doi}[1]{doi: #1}\else
  \providecommand{\doi}{doi: \begingroup \urlstyle{rm}\Url}\fi

\bibitem[He et~al.(2015)He, Zhang, Ren, and Sun]{he2015delving}
Kaiming He, Xiangyu Zhang, Shaoqing Ren, and Jian Sun.
\newblock Delving deep into rectifiers: Surpassing human-level performance on
  imagenet classification.
\newblock In \emph{ICCV}, pages 1026--1034, 2015.

\bibitem[Assael et~al.(2017)Assael, Shillingford, Whiteson, and
  De~Freitas]{assael2016lipnet}
Yannis~M Assael, Brendan Shillingford, Shimon Whiteson, and Nando De~Freitas.
\newblock Lipnet: End-to-end sentence-level lipreading.
\newblock \emph{ICLR}, 2017.

\bibitem[Peirce(1974)]{peirce1974collected}
Charles~Sanders Peirce.
\newblock \emph{{Collected papers of Charles Sanders Peirce}}, volume~5.
\newblock Harvard University Press, 1974.

\bibitem[Li et~al.(2022)Li, Wu, Fan, Mangalam, Xiong, Malik, and
  Feichtenhofer]{li2022mvitv2}
Yanghao Li, Chao-Yuan Wu, Haoqi Fan, Karttikeya Mangalam, Bo~Xiong, Jitendra
  Malik, and Christoph Feichtenhofer.
\newblock Mvitv2: Improved multiscale vision transformers for classification
  and detection.
\newblock In \emph{CVPR}, pages 4804--4814, 2022.

\bibitem[Lin et~al.(2023)Lin, Ye, Zhu, Cui, Ning, Jin, and Yuan]{lin2023video}
Bin Lin, Yang Ye, Bin Zhu, Jiaxi Cui, Munan Ning, Peng Jin, and Li~Yuan.
\newblock Video-llava: Learning united visual representation by alignment
  before projection.
\newblock \emph{arXiv preprint arXiv:2311.10122}, 2023.

\bibitem[Cherian et~al.(2022)Cherian, Hori, Marks, and Le~Roux]{cherian20222}
Anoop Cherian, Chiori Hori, Tim~K Marks, and Jonathan Le~Roux.
\newblock (2.5+ 1) d spatio-temporal scene graphs for video question answering.
\newblock In \emph{Proceedings of the AAAI Conference on Artificial
  Intelligence}, volume~36, pages 444--453, 2022.

\bibitem[Dang et~al.(2021)Dang, Le, Le, and Tran]{dang2021hierarchical}
Long~Hoang Dang, Thao~Minh Le, Vuong Le, and Truyen Tran.
\newblock Hierarchical object-oriented spatio-temporal reasoning for video
  question answering.
\newblock \emph{IJCAI}, 2021.

\bibitem[Yi et~al.(2020)Yi, Gan, Li, Kohli, Wu, Torralba, and
  Tenenbaum]{yi2020clevrer}
Kexin Yi, Chuang Gan, Yunzhu Li, Pushmeet Kohli, Jiajun Wu, Antonio Torralba,
  and Joshua~B. Tenenbaum.
\newblock {CLEVRER: CoLlision Events for Video REpresentation and Reasoning}.
\newblock In \emph{ICLR}, 2020.

\bibitem[Gao et~al.(2023)Gao, Zhou, Ji, Zhu, Yang, and Shou]{gao2023mist}
Difei Gao, Luowei Zhou, Lei Ji, Linchao Zhu, Yi~Yang, and Mike~Zheng Shou.
\newblock Mist: Multi-modal iterative spatial-temporal transformer for
  long-form video question answering.
\newblock In \emph{Proceedings of the IEEE/CVF conference on computer vision
  and pattern recognition}, pages 14773--14783, 2023.

\bibitem[Liang et~al.(2022)Liang, Wang, Zhou, and Yang]{liang2022visual}
Chen Liang, Wenguan Wang, Tianfei Zhou, and Yi~Yang.
\newblock Visual abductive reasoning.
\newblock In \emph{Proceedings of the IEEE/CVF Conference on Computer Vision
  and Pattern Recognition}, pages 15565--15575, 2022.

\bibitem[Zhao et~al.(2022)Zhao, Rao, Tang, Zhou, and Lu]{zhao2022videoabc}
Wenliang Zhao, Yongming Rao, Yansong Tang, Jie Zhou, and Jiwen Lu.
\newblock {VideoABC: A Real-World Video Dataset for Abductive Visual
  Reasoning}.
\newblock \emph{IEEE Transactions on Image Processing}, 31:\penalty0
  6048--6061, 2022.

\bibitem[Hessel et~al.(2022)Hessel, Hwang, Park, Zellers, Bhagavatula,
  Rohrbach, Saenko, and Choi]{hessel2022abduction}
Jack Hessel, Jena~D Hwang, Jae~Sung Park, Rowan Zellers, Chandra Bhagavatula,
  Anna Rohrbach, Kate Saenko, and Yejin Choi.
\newblock The abduction of sherlock holmes: A dataset for visual abductive
  reasoning.
\newblock In \emph{ECCV}, pages 558--575. Springer, 2022.

\bibitem[Suchan et~al.(2018)Suchan, Bhatt, Wa{\l}ega, and
  Schultz]{suchan2018visual}
Jakob Suchan, Mehul Bhatt, Przemys{\l}aw Wa{\l}ega, and Carl Schultz.
\newblock Visual explanation by high-level abduction: On answer-set programming
  driven reasoning about moving objects.
\newblock In \emph{AAAI}, volume~32, 2018.

\bibitem[Li et~al.(2020)Li, Torralba, Anandkumar, Fox, and Garg]{li2020causal}
Yunzhu Li, Antonio Torralba, Animashree Anandkumar, Dieter Fox, and Animesh
  Garg.
\newblock Causal discovery in physical systems from videos.
\newblock \emph{Neural Information Processing Systems (NeurIPS)}, 2020.

\bibitem[Ding et~al.(2021)Ding, Chen, Du, Luo, Tenenbaum, and
  Gan]{ding2021dynamic}
Mingyu Ding, Zhenfang Chen, Tao Du, Ping Luo, Josh Tenenbaum, and Chuang Gan.
\newblock Dynamic visual reasoning by learning differentiable physics models
  from video and language.
\newblock \emph{NeurIPS}, 34:\penalty0 887--899, 2021.

\bibitem[Ates et~al.(2022)Ates, Atesoglu, Yigit, Kesen, Kobas, Erdem, Erdem,
  Goksun, and Yuret]{ates2020craft}
Tayfun Ates, M~Samil Atesoglu, Cagatay Yigit, Ilker Kesen, Mert Kobas, Erkut
  Erdem, Aykut Erdem, Tilbe Goksun, and Deniz Yuret.
\newblock {Craft: A benchmark for causal reasoning about forces and
  interactions}.
\newblock \emph{ACL}, pages 2602--2627, 2022.

\bibitem[Locatello et~al.(2020)Locatello, Weissenborn, Unterthiner, Mahendran,
  Heigold, Uszkoreit, Dosovitskiy, and Kipf]{locatello2020object}
Francesco Locatello, Dirk Weissenborn, Thomas Unterthiner, Aravindh Mahendran,
  Georg Heigold, Jakob Uszkoreit, Alexey Dosovitskiy, and Thomas Kipf.
\newblock Object-centric learning with slot attention.
\newblock \emph{NeurIPS}, 33:\penalty0 11525--11538, 2020.

\bibitem[Girdhar and Ramanan(2019)]{girdhar2019cater}
Rohit Girdhar and Deva Ramanan.
\newblock Cater: A diagnostic dataset for compositional actions and temporal
  reasoning.
\newblock \emph{arXiv preprint arXiv:1910.04744}, 2019.

\bibitem[Mao et~al.(2022)Mao, Yang, Zhang, Goodman, and Wu]{mao2022clevrer}
Jiayuan Mao, Xuelin Yang, Xikun Zhang, Noah Goodman, and Jiajun Wu.
\newblock {CLEVRER-Humans: Describing Physical and Causal Events the Human
  Way}.
\newblock In \emph{NeurIPS Datasets and Benchmarks Track}, 2022.

\bibitem[Liu et~al.(2021)Liu, Hu, Bai, Ding, Bai, and Torr]{liu2021multi}
Xiaolong Liu, Yao Hu, Song Bai, Fei Ding, Xiang Bai, and Philip~HS Torr.
\newblock Multi-shot temporal event localization: a benchmark.
\newblock In \emph{CVPR}, pages 12596--12606, 2021.

\bibitem[Dai et~al.(2017)Dai, Singh, Zhang, Davis, and
  Qiu~Chen]{dai2017temporal}
Xiyang Dai, Bharat Singh, Guyue Zhang, Larry~S Davis, and Yan Qiu~Chen.
\newblock Temporal context network for activity localization in videos.
\newblock In \emph{ICCV}, pages 5793--5802, 2017.

\bibitem[Xia et~al.(2022)Xia, Wang, Zhou, Zheng, and Tang]{xia2022learning}
Kun Xia, Le~Wang, Sanping Zhou, Nanning Zheng, and Wei Tang.
\newblock Learning to refactor action and co-occurrence features for temporal
  action localization.
\newblock In \emph{CVPR}, pages 13884--13893, 2022.

\bibitem[Lei et~al.(2020)Lei, Yu, Berg, and Bansal]{lei2020more}
Jie Lei, Licheng Yu, Tamara~L Berg, and Mohit Bansal.
\newblock What is more likely to happen next? video-and-language future event
  prediction.
\newblock \emph{EMNLP}, pages 8769--8784, 2020.

\bibitem[Park et~al.(2020)Park, Bhagavatula, Mottaghi, Farhadi, and
  Choi]{park2020visualcomet}
Jae~Sung Park, Chandra Bhagavatula, Roozbeh Mottaghi, Ali Farhadi, and Yejin
  Choi.
\newblock Visualcomet: Reasoning about the dynamic context of a still image.
\newblock In \emph{ECCV}, pages 508--524. Springer, 2020.

\bibitem[Ji et~al.(2020)Ji, Krishna, Fei-Fei, and Niebles]{ji2020action}
Jingwei Ji, Ranjay Krishna, Li~Fei-Fei, and Juan~Carlos Niebles.
\newblock {Action genome: Actions as compositions of spatio-temporal scene
  graphs}.
\newblock In \emph{CVPR}, pages 10236--10247, 2020.

\bibitem[Sadhu et~al.(2021)Sadhu, Gupta, Yatskar, Nevatia, and
  Kembhavi]{sadhu2021visual}
Arka Sadhu, Tanmay Gupta, Mark Yatskar, Ram Nevatia, and Aniruddha Kembhavi.
\newblock Visual semantic role labeling for video understanding.
\newblock In \emph{CVPR}, pages 5589--5600, 2021.

\bibitem[Ayyubi et~al.(2022)Ayyubi, Thomas, Chum, Lokesh, Niu, Lin, Chen, Koo,
  Ray, and Chang]{ayyubi2022multimodal}
Hammad~A Ayyubi, Christopher Thomas, Lovish Chum, Rahul Lokesh, Yulei Niu,
  Xudong Lin, Long Chen, Jaywon Koo, Sounak Ray, and Shih-Fu Chang.
\newblock Multimodal event graphs: Towards event centric understanding of
  multimodal world.
\newblock \emph{arXiv preprint arXiv:2206.07207}, 2022.

\bibitem[Arnab et~al.(2021{\natexlab{a}})Arnab, Dehghani, Heigold, Sun,
  Lu{\v{c}}i{\'c}, and Schmid]{arnab2021vivit}
Anurag Arnab, Mostafa Dehghani, Georg Heigold, Chen Sun, Mario Lu{\v{c}}i{\'c},
  and Cordelia Schmid.
\newblock {Vivit: A video vision transformer}.
\newblock In \emph{ICCV}, pages 6836--6846, 2021{\natexlab{a}}.

\bibitem[Liu et~al.(2022)Liu, Ning, Cao, Wei, Zhang, Lin, and Hu]{liu2022video}
Ze~Liu, Jia Ning, Yue Cao, Yixuan Wei, Zheng Zhang, Stephen Lin, and Han Hu.
\newblock Video swin transformer.
\newblock In \emph{CVPR}, pages 3202--3211, 2022.

\bibitem[Denecker et~al.(1992)Denecker, Missiaen, and
  Bruynooghe]{denecker1992temporal}
Marc Denecker, Lode Missiaen, and Maurice Bruynooghe.
\newblock Temporal reasoning with abductive event calculus.
\newblock In \emph{ECAI}, pages 384--388. John Wiley and Sons; Chichester,
  1992.

\bibitem[Bhagavatula et~al.(2019)Bhagavatula, Bras, Malaviya, Sakaguchi,
  Holtzman, Rashkin, Downey, Yih, and Choi]{bhagavatula2019abductive}
Chandra Bhagavatula, Ronan~Le Bras, Chaitanya Malaviya, Keisuke Sakaguchi, Ari
  Holtzman, Hannah Rashkin, Doug Downey, Scott Wen-tau Yih, and Yejin Choi.
\newblock Abductive commonsense reasoning.
\newblock \emph{ICLR}, 2019.

\bibitem[Elsenbroich et~al.(2006)Elsenbroich, Kutz, and
  Sattler]{elsenbroich2006case}
Corinna Elsenbroich, Oliver Kutz, and Ulrike Sattler.
\newblock A case for abductive reasoning over ontologies.
\newblock In \emph{OWLED}, volume 216, 2006.

\bibitem[Schoenfisch et~al.(2018)Schoenfisch, Meilicke, von St{\"u}lpnagel,
  Ortmann, and Stuckenschmidt]{schoenfisch2018root}
Joerg Schoenfisch, Christian Meilicke, Janno von St{\"u}lpnagel, Jens Ortmann,
  and Heiner Stuckenschmidt.
\newblock Root cause analysis in it infrastructures using ontologies and
  abduction in markov logic networks.
\newblock \emph{Information Systems}, 74:\penalty0 103--116, 2018.

\bibitem[Pearl et~al.(2000)]{pearl2000models}
Judea Pearl et~al.
\newblock Causality: Models, reasoning and inference.
\newblock \emph{Cambridge, UK: CambridgeUniversityPress}, 19, 2000.

\bibitem[Damen et~al.(2022)Damen, Doughty, Farinella, Furnari, Kazakos, Ma,
  Moltisanti, Munro, Perrett, Price, et~al.]{damen2022rescaling}
Dima Damen, Hazel Doughty, Giovanni~Maria Farinella, Antonino Furnari,
  Evangelos Kazakos, Jian Ma, Davide Moltisanti, Jonathan Munro, Toby Perrett,
  Will Price, et~al.
\newblock Rescaling egocentric vision: Collection, pipeline and challenges for
  epic-kitchens-100.
\newblock \emph{International Journal of Computer Vision}, pages 1--23, 2022.

\bibitem[Zhong et~al.(2023)Zhong, Schneider, Voit, Stiefelhagen, and
  Beyerer]{zhong2023anticipative}
Zeyun Zhong, David Schneider, Michael Voit, Rainer Stiefelhagen, and J{\"u}rgen
  Beyerer.
\newblock Anticipative feature fusion transformer for multi-modal action
  anticipation.
\newblock In \emph{Proceedings of the IEEE/CVF Winter Conference on
  Applications of Computer Vision}, pages 6068--6077, 2023.

\bibitem[Sanh et~al.(2019)Sanh, Debut, Chaumond, and Wolf]{sanh2019distilbert}
Victor Sanh, Lysandre Debut, Julien Chaumond, and Thomas Wolf.
\newblock Distilbert, a distilled version of bert: smaller, faster, cheaper and
  lighter.
\newblock \emph{NeurIPS}, 2019.

\bibitem[Zhang et~al.(2013)Zhang, Zhang, Swears, Larios, Wang, and
  Ji]{zhang2013modeling}
Yongmian Zhang, Yifan Zhang, Eran Swears, Natalia Larios, Ziheng Wang, and
  Qiang Ji.
\newblock Modeling temporal interactions with interval temporal bayesian
  networks for complex activity recognition.
\newblock \emph{TPAMI}, 35\penalty0 (10):\penalty0 2468--2483, 2013.

\bibitem[Allen(1983)]{allen1983maintaining}
James~F Allen.
\newblock Maintaining knowledge about temporal intervals.
\newblock \emph{Communications of the ACM}, 26\penalty0 (11):\penalty0
  832--843, 1983.

\bibitem[Arnab et~al.(2021{\natexlab{b}})Arnab, Sun, and
  Schmid]{arnab2021unified}
Anurag Arnab, Chen Sun, and Cordelia Schmid.
\newblock Unified graph structured models for video understanding.
\newblock In \emph{ICCV}, pages 8117--8126, 2021{\natexlab{b}}.

\bibitem[Mao et~al.(2018)Mao, Wu, Xue, and Zhang]{mao2018hierarchical}
Feng Mao, Xiang Wu, Hui Xue, and Rong Zhang.
\newblock Hierarchical video frame sequence representation with deep
  convolutional graph network.
\newblock In \emph{ECCV workshops}, pages 0--0, 2018.

\bibitem[Wang et~al.(2019)Wang, Lu, Shen, Crandall, and Shao]{wang2019zero}
Wenguan Wang, Xiankai Lu, Jianbing Shen, David~J Crandall, and Ling Shao.
\newblock Zero-shot video object segmentation via attentive graph neural
  networks.
\newblock In \emph{ICCV}, pages 9236--9245, 2019.

\bibitem[Feng et~al.(2022)Feng, Chen, Li, Sarkar, and Zhang]{feng2022powerful}
Jiarui Feng, Yixin Chen, Fuhai Li, Anindya Sarkar, and Muhan Zhang.
\newblock How powerful are k-hop message passing graph neural networks.
\newblock \emph{NeurIPS}, 2022.

\bibitem[Hochreiter and Schmidhuber(1997)]{hochreiter1997long}
Sepp Hochreiter and J{\"u}rgen Schmidhuber.
\newblock Long short-term memory.
\newblock \emph{Neural computation}, 9\penalty0 (8):\penalty0 1735--1780, 1997.

\bibitem[Graves and Schmidhuber(2005)]{graves2005framewise}
Alex Graves and J{\"u}rgen Schmidhuber.
\newblock Framewise phoneme classification with bidirectional lstm and other
  neural network architectures.
\newblock \emph{Neural networks}, 18\penalty0 (5-6):\penalty0 602--610, 2005.

\bibitem[Vaswani et~al.(2017)Vaswani, Shazeer, Parmar, Uszkoreit, Jones, Gomez,
  Kaiser, and Polosukhin]{vaswani2017attention}
Ashish Vaswani, Noam Shazeer, Niki Parmar, Jakob Uszkoreit, Llion Jones,
  Aidan~N Gomez, {\L}ukasz Kaiser, and Illia Polosukhin.
\newblock Attention is all you need.
\newblock \emph{NIPS}, 30, 2017.

\bibitem[Kay et~al.(2017)Kay, Carreira, Simonyan, Zhang, Hillier,
  Vijayanarasimhan, Viola, Green, Back, Natsev, et~al.]{kay2017kinetics}
Will Kay, Joao Carreira, Karen Simonyan, Brian Zhang, Chloe Hillier, Sudheendra
  Vijayanarasimhan, Fabio Viola, Tim Green, Trevor Back, Paul Natsev, et~al.
\newblock The kinetics human action video dataset.
\newblock \emph{arXiv preprint arXiv:1705.06950}, 2017.

\bibitem[Kingma and Ba(2014)]{kingma2014adam}
Diederik Kingma and Jimmy Ba.
\newblock Adam: A method for stochastic optimization.
\newblock \emph{ICLR}, 2014.

\end{thebibliography}
\clearpage}{\small\par}

\appendix
\part*{Supplementary material}
\section*{Introduction}

In this supplementary material, we provide further details on the
implementation of our proposed method $\ModelName$ and the baselines
discussed in the Experiment section in the main paper and additional
analyses. This includes: 
\begin{itemize}
\item Additional details of $\ModelName$
\item Implementation details 
\item Some qualitative results on the $\Dataset$ dataset
\end{itemize}

\section{Additional Details of $\protect\ModelName$}

\subsection{Event temporal relations}

\begin{table}[h]
\begin{centering}
\begin{tabular}{|>{\centering}p{0.75cm}|>{\centering}p{0.75cm}|>{\centering}p{0.75cm}|>{\centering}p{0.75cm}|>{\centering}p{3.2cm}|}
\hline 
{\scriptsize{}$t_{j}^{s}-t_{i}^{s}$} & {\scriptsize{}$t_{j}^{e}-t_{i}^{e}$} & {\scriptsize{}$t_{j}^{s}-t_{i}^{e}$} & {\scriptsize{}$t_{j}^{e}-t_{i}^{s}$} & {\scriptsize{}Interpretation}\tabularnewline
\hline 
\hline 
{\scriptsize{}+} & {\scriptsize{}+} & {\scriptsize{}+} & {\scriptsize{}+} & {\scriptsize{}$e_{i}$ before $e_{j}$ ($e_{i}\rightarrow e_{j}$)}\tabularnewline
\hline 
{\scriptsize{}-} & {\scriptsize{}-} & {\scriptsize{}-} & {\scriptsize{}-} & {\scriptsize{}$e_{j}$ before $e_{i}$ ($e_{i}\leftarrow e_{j}$)}\tabularnewline
\hline 
{\scriptsize{}-} & {\scriptsize{}+} & {\scriptsize{}+} & {\scriptsize{}-} & {\scriptsize{}$e_{i}$ during $e_{j}$ ($e_{i}\rightarrow e_{j}$)}\tabularnewline
\hline 
{\scriptsize{}+} & {\scriptsize{}-} & {\scriptsize{}-} & {\scriptsize{}+} & {\scriptsize{}$e_{j}$ during $e_{i}$ ($e_{i}\leftarrow e_{j}$)}\tabularnewline
\hline 
{\scriptsize{}+} & {\scriptsize{}+} & {\scriptsize{}+} & {\scriptsize{}-} & {\scriptsize{}$e_{i}$ overlaps $e_{j}$ ($e_{i}\rightarrow e_{j}$)}\tabularnewline
\hline 
{\scriptsize{}-} & {\scriptsize{}-} & {\scriptsize{}+} & {\scriptsize{}-} & {\scriptsize{}$e_{j}$ overlaps $e_{i}$ ($e_{i}\leftarrow e_{j}$)}\tabularnewline
\hline 
{\scriptsize{}+} & {\scriptsize{}+} & {\scriptsize{}+} & {\scriptsize{}0} & {\scriptsize{}$e_{i}$ meets $e_{p}$ ($e_{i}\rightarrow e_{j}$)}\tabularnewline
\hline 
{\scriptsize{}-} & {\scriptsize{}-} & {\scriptsize{}0} & {\scriptsize{}-} & {\scriptsize{}$e_{j}$ meets $e_{i}$ ($e_{i}\leftarrow e_{j}$)}\tabularnewline
\hline 
{\scriptsize{}0} & {\scriptsize{}+} & {\scriptsize{}-} & {\scriptsize{}+} & {\scriptsize{}$e_{i}$ starts $e_{p}$ ($e_{i}\rightarrow e_{j}$)}\tabularnewline
\hline 
{\scriptsize{}0} & {\scriptsize{}-} & {\scriptsize{}-} & {\scriptsize{}+} & {\scriptsize{}$e_{p}$ starts $e_{i}$ ($e_{i}\leftarrow e_{j}$)}\tabularnewline
\hline 
{\scriptsize{}-} & {\scriptsize{}0} & {\scriptsize{}-} & {\scriptsize{}+} & {\scriptsize{}$e_{i}$ finishes $e_{j}$ ($e_{i}\rightarrow e_{j}$)}\tabularnewline
\hline 
{\scriptsize{}+} & {\scriptsize{}0} & {\scriptsize{}-} & {\scriptsize{}+} & {\scriptsize{}$e_{j}$ finishes $e_{i}$ ($e_{i}\leftarrow e_{j}$)}\tabularnewline
\hline 
{\scriptsize{}0} & {\scriptsize{}0} & {\scriptsize{}any} & {\scriptsize{}any} & {\scriptsize{}$e_{i}$ equals $e_{j}$ ($e_{i}\rightarrow e_{j}$)}\tabularnewline
\hline 
\end{tabular}\medskip{}
\par\end{centering}
\caption{Identifying the temporal order between event $e_{i}$ and $e_{p}$
based on Allen's atomic interval temporal relations.\label{tab:allen_algebra}}
\end{table}

As mentioned in the main paper, we use Allen's interval algebra \cite{allen1983maintaining}
to determine the temporal relations between a pair of temporal events
$e_{i}$ and $e_{j}$. Specifically, given the temporal distance between
the two events as in the Eq.~3 in the main paper, their order in
time is determined using in Table~\ref{tab:allen_algebra}. As can
be seen, these orders are mainly based on the events' start times.
If the two events have the same start time, their end times will be
considered.

\subsection{Event semantic relations}

The distance-based penalty factor of temporal distance $\gamma_{ip}$
in Eq.~5 of the main paper is computed by: 

\begin{equation}
\gamma_{ij}=\text{exp}\left(-\beta*d_{ij}\right),\,\text{where}
\end{equation}
 
\begin{equation}
d_{ij}\left(e_{i},e_{j}\right)_{i<j}=\sqrt{\left(ts_{j}-ts_{i}\right)^{2}+\left(te_{j}-te_{i}\right)^{2}}
\end{equation}
is the Euclidean distance between events $e_{i}$ and $e_{j}$; $\beta$
is a learnable positive decay factor that compensates for the uncertainty
in calculating the distance $d_{ij}$. The more distant the two events
are in time, the weaker their relation is.

\section{Implementation Details}

\subsection{CERN}

We implemented our proposed method $\ModelName$ using DGL 0.9.1 and
Pytorch 1.11.0 or later. We have experimented with up to 4 layers
of message passing in which we found slight improvements when increasing
the number of layers as reported in the Experiment section in the
main paper. Our proposed model is optimized using Adam optimizer \cite{kingma2014adam}
with initial learning rate is set to $10^{-4}$. The learning rate
gradually decreases at every epoch. We train a total $100$ epochs
with a batch size of $128$ which takes approximately $18$ hours
on a single GPU NVIDIA A100 to finish the training on the full $\Dataset$
dataset.

\subsection{Baselines}

This section provides details of the baselines used in the Experiment
section, which include:

\textbf{First collision:} This experiment aims to verify whether it
is sufficient to always set the first collision detected in video
input as the trigger event, regardless of the target event considered.

\textbf{Node embeddings:} Directly use node embeddings of the events
by using a linear neural network layer. This experiment does not consider
video structure and relationships of any kinds between the nodes.

\textbf{LSTM \cite{hochreiter1997long}: }We first sort all the events
localized from an input video by their occurrence in time. We then
simply treat them as a sequence and use LSTM networks to model the
temporal relationships between the events. The output hidden states
of LSTM are then used as the latent representation $h_{i}$ of the
events in Eq.~(12) in the main paper.

\textbf{BiLSTM \cite{graves2005framewise}:} Instead of using LSTM,
we use a bidirectional LSTM variant to model the sequential relationships
between events. The objective is to take advantage of signals propagating
both forward and backward in time.

\textbf{Transformer \cite{vaswani2017attention}: }Similarly, we also
sort all the events by their occurrence times to create a sequence
of events. We implement a vanilla version of Transformer with 6 layers
of self-attention to refine the representations of the events by taking
into account their relationships with all other events in the sequence.
We use the same parameters as suggested in the original paper in our
implementation.

\textbf{Video-LLaVA \cite{lin2023video}}: To evaluate whether large
video-language models can comprehend the causal relationships between
video events, we conduct experiments with Video-LLaVA on the EpicKitchen-AR
dataset using two settings: zero-shot setting and in-context setting. 

In the zero-shot setting, Video-LLaVA is instructed to respond to
a templated question where it must identify the trigger event for
a target event from a list of premise events. Careful instructions
are provided to ensure consistent responses from Video-LLaVA. We have
tried different prompts and reported the one with the best results.

In the in-context learning setting, for each sample in the test set
of the EpicKitchen-AR, we provide 2 samples with grouthtruth answers
taken from the train set. These examples serve as context to guide
Video-LLaVA's responses. The same instruction format as the zero-shot
setting is used in this experiment. Specifically, the following prompt
is given to Video-LLaVA :

\emph{`` USER: \textless video\textgreater Given a sequence of
premise events include \{premise\_events\}, what is the cause event
of the event \{target\_event\}? Choose a correct answer within the
premise events. ASSISTANT:''}

All of these baselines are implemented using Pytorch 1.11.0 or later
with similar hyper-parameters as our proposed model $\ModelName$
for fair comparisons.

\section{Analysis}

\begin{figure*}
\begin{centering}
\includegraphics[width=0.95\columnwidth]{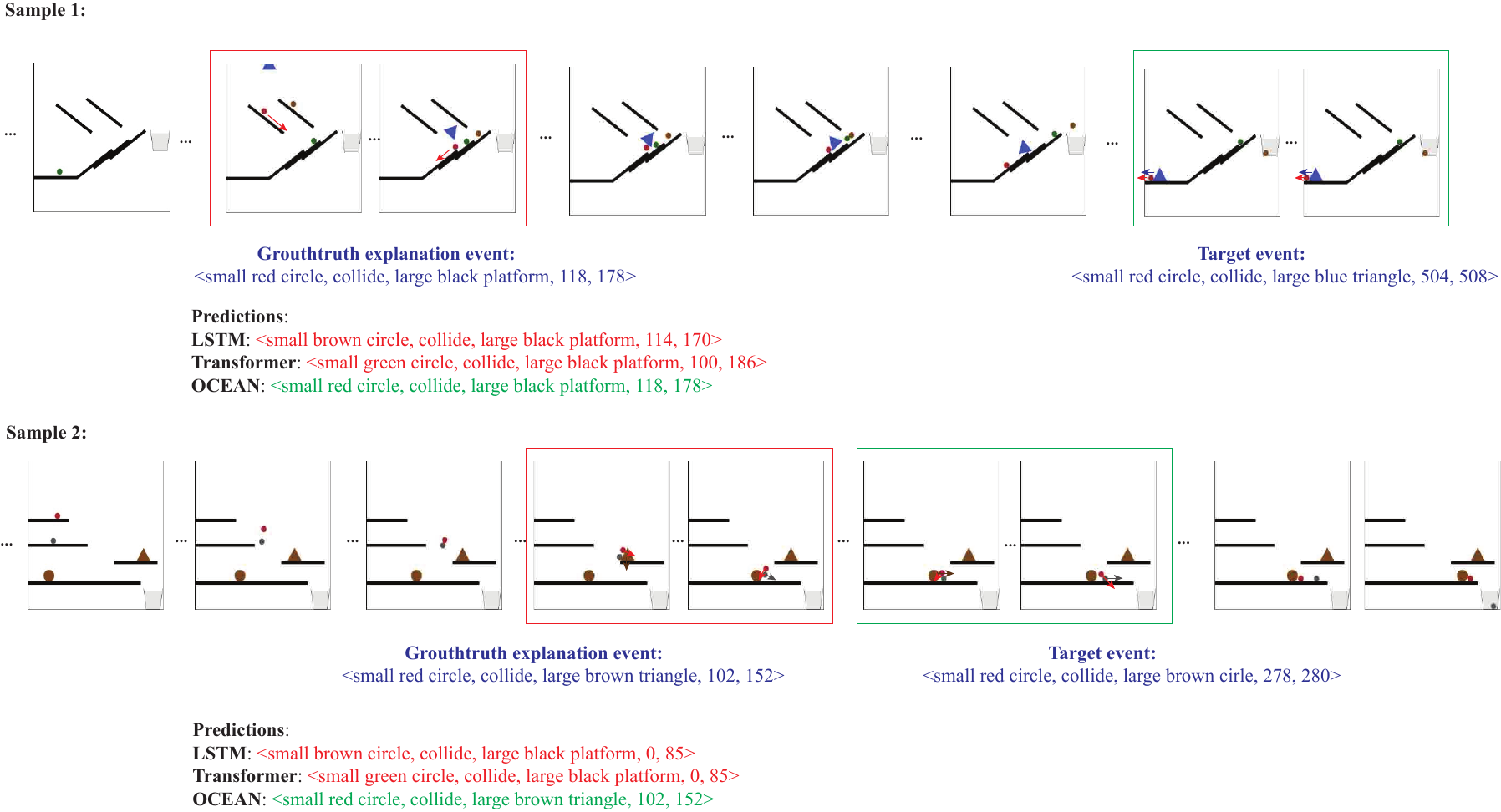}
\par\end{centering}
\caption{Qualitative examples demonstrating that sequential models struggle
to identify correct trigger events while $\protect\ModelName$ handles
successfully. Sequential models tend to predict events that occur
early in time as the trigger events and often incapable of tracing
back events associated with a co-referenced object between the target
and trigger event. The graph structure learned by $\protect\ModelName$
has potential to facilitate the propagation of information along chains
of events associated with dynamic objects, thereby benefiting the
learning. Colored arrows indicate the direction of the movement of
the corresponding objects. Best viewed in color.\label{fig:Qualitative-type-1}}
\end{figure*}

Figure \ref{fig:Qualitative-type-1} presents typical examples where
the graph structure modeled by CERN helps trace back events associated
with a co-referenced object between target-trigger events to identify
the correct trigger events. In contrast, LSTM and Transformer struggle
to keep track of these event sequences due to their assumptions on
the sequential relationships between events. Furthermore, we also
observe that LSTM and Transformer often make similar mistakes in these
examples as they tend to select events that occur at beginning of
a video input as trigger events, thus fail to identify trigger events
that locate at the middle of the video.

\begin{figure*}
\begin{centering}
\includegraphics[width=0.95\columnwidth]{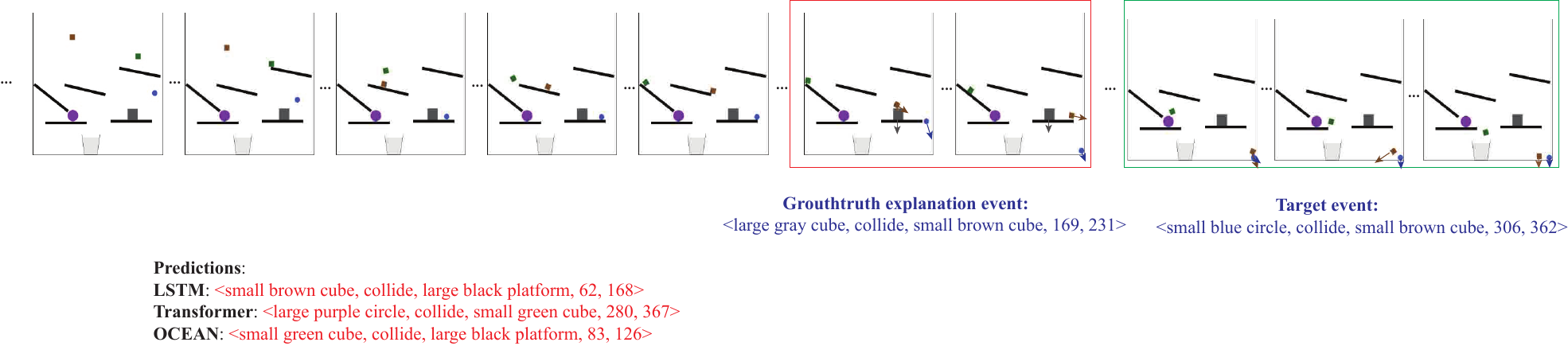}
\par\end{centering}
\caption{A showcase for the challenges of our $\protect\Dataset$ dataset that
all the studied methods struggle. This is often the case when there
are multiple concurrent events occurring within the times of the trigger
event. Best viewed in color.\label{fig:All-wrong}}
\end{figure*}

However, we also showcase an example where all the studied methods
struggle in Figure \ref{fig:All-wrong}. This often occurs when there
are multiple concurrent events happen within a scene during the life
of the trigger event. The inability of these models to handle such
scenarios suggest a need for a novel approach that goes beyond the
common feature association between events.

\end{document}